%% file: acl2023.tex
\pdfoutput=1

\documentclass[11pt]{article}

\usepackage[]{ACL2023}

\usepackage{times}
\usepackage{latexsym}

\usepackage[T1]{fontenc}

\usepackage[utf8]{inputenc}

\usepackage{microtype}

\usepackage{inconsolata}

\usepackage[ruled, vlined, linesnumbered]{algorithm2e}
\usepackage{bm}
\usepackage{graphicx}
\usepackage{amsmath}
\usepackage{multirow}
\usepackage{cleveref}
\usepackage{xspace}
\usepackage{xcolor}
\usepackage{enumitem}
\usepackage{placeins}
\usepackage{soul}
\usepackage{bbm}
\usepackage{booktabs}
\usepackage{hyperref}
\usepackage{subcaption}
\usepackage{pgfplots}
\usepackage{amssymb}
\DeclareUnicodeCharacter{2212}{−}
\usepgfplotslibrary{groupplots,dateplot}
\usetikzlibrary{patterns,shapes.arrows}
\pgfplotsset{compat=newest}

\newcommand{\ours}{{MetaEvent}\xspace}

\title{Zero- and Few-Shot Event Detection via Prompt-Based Meta Learning}

\author{{Zhenrui Yue \quad Huimin Zeng \quad Mengfei Lan \quad Heng Ji \quad Dong Wang} \\
  University of Illinois Urbana-Champaign \\
  \texttt{\{zhenrui3, huiminz3, mlan3, hengji, dwang24\}@illinois.edu}}

\begin{document}
\maketitle
\begin{abstract}
With emerging online topics as a source for numerous new events, detecting unseen / rare event types presents an elusive challenge for existing event detection methods, where only limited data access is provided for training. To address the data scarcity problem in event detection, we propose \ours, a meta learning-based framework for zero- and few-shot event detection. Specifically, we sample training tasks from existing event types and perform meta training to search for optimal parameters that quickly adapt to unseen tasks. In our framework, we propose to use the cloze-based prompt and a trigger-aware soft verbalizer to efficiently project output to unseen event types. Moreover, we design a contrastive meta objective based on maximum mean discrepancy (MMD) to learn class-separating features. As such, the proposed \ours can perform zero-shot event detection by mapping features to event types without any prior knowledge. In our experiments, we demonstrate the effectiveness of \ours in both zero-shot and few-shot scenarios, where the proposed method achieves state-of-the-art performance in extensive experiments on benchmark datasets FewEvent and MAVEN.
\end{abstract}

\input{1_intro}
\input{2_related.tex}
\input{3_method}
\input{4_experiment}
\input{5_conclusion}




\bibliography{anthology, custom}
\bibliographystyle{acl_natbib}

\clearpage
\appendix
\input{6_appendix.tex}

\end{document}

%% file: 1_intro.tex
\section{Introduction}
\label{sec:intro}

\begin{figure}[t]
    \centering
    \includegraphics[trim=7.4cm 4.2cm 7.7cm 3.8cm, clip, width=1.0\linewidth]{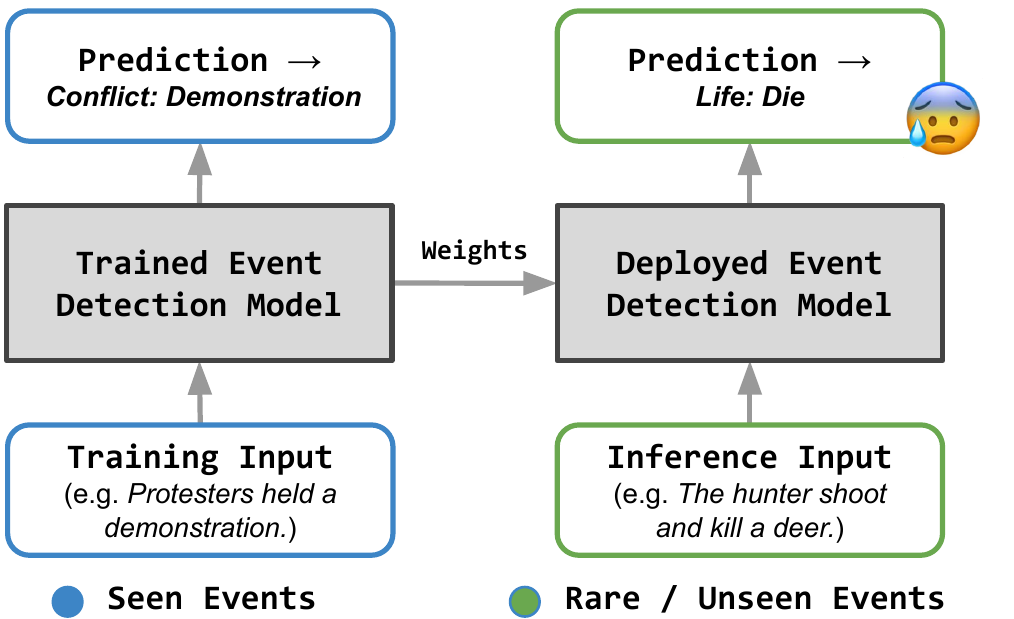}
    \caption{Existing event detection model fails to detect rare / unseen events (green input) upon deployment.} 
    \label{fig:intro}
    \vspace{-10pt}
\end{figure}

Event detection tasks have experienced significant improvements thanks to the recent efforts in developing language-based methods~\cite{lu-etal-2021-text2event, pouran-ben-veyseh-etal-2021-unleash}. One of such methods is pretrained large language models, which can be fine-tuned for detecting events upon input context~\cite{liu2019roberta, cong-etal-2021-shot}. However, the detection of unseen or rare events remains a challenge for existing methods, as large amounts of annotated data are required for training in a supervised fashion~\cite{shang2022sat, zhang2022active}. For instance, existing models often fail to detect unseen event types due to the lack of domain knowledge, as demonstrated in \Cref{fig:intro}.

To detect unseen event types (i.e., zero-shot learning), existing methods leverage external knowledge or contrastive learning to build class-separating features~\cite{lyu-etal-2021-zero, zhang2022efficient, zhang-etal-2022-zero}. Similarly, it is possible to leverage limited annotated examples (i.e., few-shot learning) for event detection. For instance, prototypical features can be constructed to match event types upon inference~\cite{deng2020meta, cong-etal-2021-shot}. Prompt-based methods align with the pretraining objective of language models to improve detection performance~\cite{li-etal-2022-kipt, li2022piled}. Specifically, a cloze-based prompt template (e.g., \texttt{A <mask> event}) is incorporated as part of the input, and the prediction can be obtained by decoding the \texttt{<mask>} prediction. As such, prompt-based methods exploit the masked language modeling (MLM) pretraining task by constructing similar input examples.

Nevertheless, no existing method is designed for both zero-shot and few-shot event detection, as it is a non-trivial problem to combine both settings under a unified framework. Previous efforts either address the zero-shot or the few-shot setting, yet detecting both unseen and rare events can be helpful in various scenarios (e.g., detecting emergency events online), which renders current approaches less effective in real-world applications. Additionally, event detection comprises of two subtasks: trigger identification and classification. Some methods provide an incomplete solution by solely performing the classification task, while many other approaches execute both tasks at the cost of reduced performance~\cite{schick-schutze-2021-exploiting, li2022piled}. This is because existing methods heavily rely on trigger words for classification, which causes drastic performance drops in the case of trigger mismatch or ambiguity~\cite{ding-etal-2019-event}.

In this paper, we propose a meta learning framework \ours for zero- and few-shot event detection. We consider both settings and optimize language models to identify unseen / rare events via a unified meta learning framework. That is, given zero / limited number of annotated examples per event type, our objective is to maximize the model performance in detecting such events. To this end, we develop a solution to integrate the trigger identification and classification subtasks for efficient forward passing in meta training. Moreover, we design a trigger-aware soft verbalizer to identify event types in our prompt-based event detection model. For optimization, we propose a meta objective function based on contrastive loss to learn generalizable and class-separating event features. In training, the model is first updated and evaluated upon the sampled zero- or few-shot tasks, then the meta loss is computed to derive gradients w.r.t. the initial parameters, such that the updated model learns to generalize to the target tasks even without labeled examples. In other words, \ours learns from seen tasks, yet with the objective to generalize in rare and unseen scenarios. Therefore, the resulting model can optimally adapt to the target task upon deployment. We demonstrate the effectiveness of \ours by evaluating zero- and few-shot event detection tasks on benchmark datasets, where \ours consistently outperforms state-of-the-art methods with considerable improvements.

We summarize our contributions as follows\footnote{We adopt publicly available datasets in the experiments and release our implementation at https://github.com/Yueeeeeeee/MetaEvent.}:
\begin{enumerate}
\item To the best of our knowledge, we are the first to propose a unified meta learning framework for both zero- and few-shot event detection. \ours is designed to exploit prompt tuning and contrastive learning for quick adaptation to unseen tasks.
\item We propose an integrated trigger-aware model in \ours for efficient meta training. In particular, our trigger-aware soft verbalizer leverages both the prompt output and attentive trigger features to identify event types.
\item We design a novel contrastive loss as the meta objective of \ours. Our contrastive loss encourages class-separating and generalizable features to improve event matching in both zero- and few-shot event detection.
\item We demonstrate the effectiveness of \ours with extensive experiments, where \ours outperforms state-of-the-art baseline methods with considerable improvements in both zero- and few-shot event detection.
\end{enumerate}

%% file: 2_related.tex
\section{Related Work}

\subsection{Event Detection}
Event detection refers to the task of classifying event types upon input text. While event detection has achieved progress under the supervised training paradigm~\cite{ji2008refining,LinACL2020,wadden-etal-2019-entity, liu-etal-2020-event, du-cardie-2020-event, lu-etal-2021-text2event, liu-etal-2022-dynamic}, zero- and few-shot classification remains a challenge due to the lack of prior knowledge and annotated examples. For the zero-shot setting, relevant literature focuses on predefined event knowledge or heuristics to classify unseen events~\cite{Huang2018,huang-ji-2020-semi, lyu-etal-2021-zero, zhang-etal-2021-zero,weaksupervision2022,YuEACL2023b,glen2023}. Similarly, zero-shot contrastive learning requires unlabeled examples in training to learn class-separating features~\cite{zhang-etal-2022-zero}. Under the few-shot setting, prototypical networks and prompt-based tuning improve detection performance via prototype matching and alignment to language pretraining~\cite{deng2020meta, cong-etal-2021-shot, schick-schutze-2021-exploiting, lai-etal-2021-learning, li2022piled}. Overall, existing approaches focus on either zero- or few-shot event detection without considering a unified framework for both settings. Moreover, current zero-shot methods require additional resources for training, making such methods less realistic in real-world event detection. As such, we propose a meta learning framework \ours for both zero- and few-shot event detection, where neither prior knowledge nor unlabeled examples are required to detect unseen events.

\subsection{Prompt Learning}
Prompt learning uses a predefined template with slots (i.e., \texttt{A <mask> event}) to instruct language models on the desired task, where the slot prediction is used to derive the final output~\cite{brown2020language, liu2021pre}. By leveraging the pretraining objective and designing prompt templates, pretrained large language models can be adapted for zero- or few-shot downstream tasks~\cite{houlsby2019parameter, raffel2020exploring}. Soft and multi-task prompts further improve the zero- and few-shot performance of language models on unseen tasks~\cite{lester-etal-2021-power, sanh2021multitask}. Cloze-based prompts are proposed for the event detection task, where the predicted output can be mapped to the event types using verbalizer or mapping heuristics~\cite{schick-schutze-2021-exploiting, li2022piled, zhang-etal-2022-zero}. Nevertheless, previous prompt methods adopt the inefficient two-step paradigm (i.e., trigger identification and classification) and are not designed for both zero- and few-shot event detection. Therefore, we integrate both steps with a trigger-aware soft verbalizer for efficient forward passing in meta training. Moreover, we consider both zero- and few-shot scenarios with our prompt-based meta learning framework \ours. By leveraging the proposed components, our approach demonstrates considerable improvements compared to existing methods.

\subsection{Meta Learning}
Meta learning (i.e., learning to learn) has demonstrated superiority in few-shot learning~\cite{finn2017model, nichol2018first, rajeswaran2019meta}. Existing methods learn class-wise features or prototypes in the metric space for quick adaptation to few-shot tasks~\cite{vinyals2016matching, snell2017prototypical, sung2018learning}. Model-agnostic meta learning (MAML) leverages the second-order optimization to find the optimal initial parameters for a new task~\cite{finn2017model}. Approximation methods of second-order MAML demonstrates comparable performance while requiring significantly reduced computational resources~\cite{finn2017model, nichol2018first, rajeswaran2019meta}. Meta learning has also been applied to tasks like online learning, domain adaptation and multi-task learning~\cite{finn2019online, li2020online, wang2021bridging}. For event detection, meta learning is proposed to improve the performance on small-size data via memory-based prototypical networks and external knowledge~\cite{deng2020meta, shen-etal-2021-adaptive}.

To the best of our knowledge, meta learning-based methods for both zero- and few-shot event detection is not studied in current literature. However, detecting unseen and rare events is necessary in many applications. An example can be detecting emergency events on social media, where both unseen and rare events can be present (e.g., pandemic, safety alert etc.). Therefore, we propose \ours: a meta learning framework for zero- and few-shot event detection. \ours exploits seen events via trigger-aware prompting and a carefully designed meta objective, and thereby improving the zero- and few-shot performance with class-separating and generalizable features.

\begin{figure*}[t]
    \centering
    \includegraphics[trim=1cm 2.5cm 2cm 2.5cm, clip, width=1.0\linewidth]{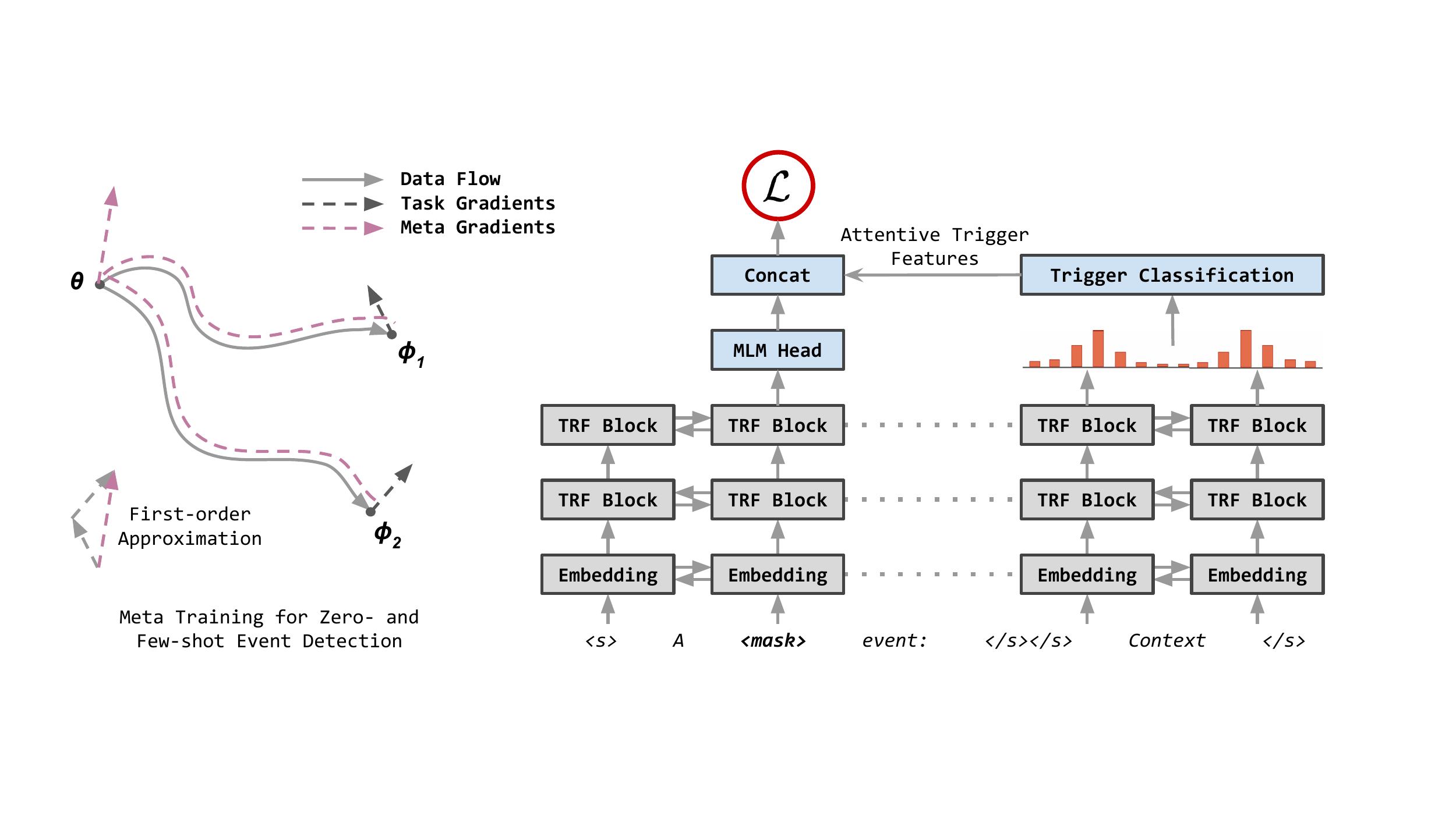}
    \caption{The proposed \ours. The left subfigure illustrates the optimization process w.r.t. the initial parameter set $\bm{\theta}$ with meta learning, and the right subfigure describes the proposed event detection model in \ours.}
    \label{fig:method}
    \vspace{-10pt}
\end{figure*}

%% file: 3_method.tex
\section{Preliminary}
We consider the following event detection problem setup, where \emph{$N$-way $K$-shot} examples are available for training each task ($K$ is 0 for zero-shot setting). Our goal is to train a model $\bm{f}$ that maximizes the performance in unseen tasks.

\noindent
\textbf{Problem}: Our research focuses on zero- and few-shot event detection based on a collection of tasks $\{ \bm{\mathcal{T}}_{i} \}_{i=1}^{M}$. For each task $\bm{\mathcal{T}}_{i}$, a $N$-way $K$-shot training set and a held-out evaluation set are provided (i.e., $\bm{\mathcal{D}}_{i}^{\mathrm{train}}, \bm{\mathcal{D}}_{i}^{\mathrm{test}} \in \bm{\mathcal{T}}_{i}$). The training of \ours is two-fold: (1)~an initial model is updated using the training sets in each sampled task to achieve local convergence (i.e., inner-loop optimization); (2)~the updated models are used to compute a meta loss on the corresponding evaluation sets, followed by deriving the gradients w.r.t. the initial model using our meta learning algorithm (i.e., outer-loop optimization). In particular, the input for each task $\bm{\mathcal{T}}_{i}$ include:
\begin{itemize}[leftmargin=10pt]
    \item \emph{Training set}: $\bm{\mathcal{D}}_{i}^{\mathrm{train}}$ contains $K$ examples for each of the $N$ classes. An example comprises of context $\bm{x}_{c}^{(j)}$, trigger $\bm{x}_{c}^{(j)}$ and label $y^{(j)}$ (i.e., $\bm{\mathcal{D}}_{i}^{\mathrm{train}} = \{ (\bm{x}_{c}^{(j)}, \bm{x}_{t}^{(j)}, y^{(j)}) \}_{j=1}^{N*K}$). In the few-shot setting, $\bm{\mathcal{D}}_{i}^{\mathrm{train}}$ is used to tune the model, while $\bm{\mathcal{D}}_{i}^{\mathrm{train}}$ is an empty set for zero-shot setting. Training set is also known as support set. 
    \item \emph{Evaluation set}: Similarly, a held-out evaluation set $\bm{\mathcal{D}}_{i}^{\mathrm{test}}$ from the same input \& label space is used to compute the meta loss in training. Upon deployment, the model can be updated with $\bm{\mathcal{D}}_{i}^{\mathrm{train}}$ and should perform well on $\bm{\mathcal{D}}_{i}^{\mathrm{test}}$. Evaluation set is often referred to as query set. 
\end{itemize}

\noindent
\textbf{Pipeline}: We are interested in learning an encoder model $\bm{f}$ parameterized by $\bm{\theta}$. Conventional event detection methods compute trigger as an intermediate variable, followed by the event type classification. As such, we formulate $\bm{f}_{\bm{\theta}}$ with contexts as input and event features as output, followed by some classifier \texttt{CLF} to map the features to the desired event type (i.e., $y = \texttt{CLF}(\bm{f}_{\bm{\theta}}(\bm{x}_{c}))$). Our goal is to find the optimal parameter set $\bm{\theta}$ that quickly adapts to an unseen task $\bm{\mathcal{T}}_{i}$ using $\bm{\mathcal{D}}_{i}^{\mathrm{train}}$ and maximizes the performance on evaluation set $\bm{\mathcal{D}}_{i}^{\mathrm{test}}$. Mathematically, this is formulated as optimization of $\bm{\theta}$ over a collection of $M$ evaluation tasks:
\begin{equation}
    \label{eq:objective}
    \min_{\substack{\bm{\theta}}} \frac{1}{M} \sum_{i=1}^{M} \mathcal{L}(\mathcal{A}lg(\bm{\theta}, \bm{\mathcal{D}}_{i}^{\mathrm{train}}), \bm{\mathcal{D}}_{i}^{\mathrm{test}}),
\end{equation}
where $\mathcal{L}$ represents the loss and $\mathcal{A}lg$ represents the gradient descent optimization algorithm.

\section{Methodology}

\subsection{Model Design}
\label{sec:model_design}
To efficiently perform prompt-based meta training for event detection, we design a one-step model that integrates the trigger identification and classification stages. This is because a two-step approach (e.g., P4E~\cite{li2022piled}) requires extensive computational resources to obtain the gradients in the inner- and outer-loop optimization in \ours. Consequently, we design an efficient model (as illustrated in \Cref{fig:method}) that integrates attentive trigger features to avoid additional forward passes.

Different from existing trigger-aware methods~\cite{ding-etal-2019-event}, the proposed model innovatively uses both attentive trigger features and prompt output to predict the event types. The attentive trigger features $\bm{t}$ can be computed using an integrated trigger classifier and attention weights from the pretrained language model. Specifically, a trigger classifier is trained upon each token features to perform binary classification (i.e., whether input token is trigger token). In inference, the classifier predicts the probabilities $\bm{p}$ of input tokens being classified as trigger words. To better estimate the importance of predicted trigger tokens, we design an attentive reweighting strategy to select informative trigger features from the input context. The idea behind our attentive reweighting strategy is to leverage attention scores from the model to select more relevant features. The attention scores reveal different importance weights of the context tokens and thus, can be used to compute `soft' trigger features based on the semantics. Formally, our attentive reweighting strategy computes weights $\bm{w} \in \bm{R}^{L_{c}}$ using trigger probabilities $\bm{p}$ and attention scores $A \in \bm{R}^{H \times L_{c} \times L_{c}}$ of the context span from the last transformer layer. The weight of the $i$-th token $w_{i}$ in $\bm{w}$ is computed via
\begin{equation}
    w_{i} = \sigma\bigg(\bm{p} \odot \frac{1}{H}\sum_{j}^{H} \bigg(\sum_{k}^{L_{c}} A_{j, k} \bigg) \bigg)_{i},
\end{equation}
where $H$ is the number of attention heads, $L_{c}$ is the context length, $\odot$ and $\sigma$ denote elementwise product and softmax functions. Base on input $\bm{x}_{c}$ and $\bm{w}$, the attentive trigger features $\bm{t}$ can be computed as the weighted sum of token features, i.e.,
\begin{equation}
    \bm{t} = \sum_{i=1}^{L_{c}} w_{i} \bm{f}_{\bm{\theta}}(\bm{x}_{c})_{i}.
\end{equation}

For the event classification, we design a prompt-based paradigm using a predefined prompt and a trigger-aware soft verbalizer. Specifically, we preprocess the input context by prepending the prompt `\texttt{A <mask> event}' to transform the prediction into a masked language modeling (MLM) task. The pretrained encoder model and MLM head fill the \texttt{<mask>} position with a probability distribution $\bm{v}$ over all tokens. Then, our trigger-aware soft verbalizer maps the predicted distribution $\bm{v}$ to an output event type. Unlike predefined verbalizer functions~\cite{li2022piled}, we design a learnable verbalizer based on MLM predictions $\bm{v}$ and attentive trigger features $\bm{t}$. For the $N$-way few-shot setting, the trigger-aware soft verbalizer with weights $\bm{W} \in \bm{R}^{(|\bm{v}|+|\bm{t}|) \times N}$ and bias $\bm{b} \in \bm{R}^{N}$ predicts the output label with GELU activation via
\begin{equation}
    \hat{y} = \arg\max (\mathrm{GELU}([\bm{v};\bm{t}])\bm{W} + \bm{b}).
\end{equation}
For zero-shot event detection, we use the concatenated features $[\bm{v};\bm{t}]$ to project input to unseen event types via the Hungarian algorithm.

\subsection{Meta Training}
Provided with the training and evaluation sets from sampled tasks, we present the formulation of our \ours and our methods for the zero- and few-shot training. The designed framework leverages meta training to search for optimal parameters $\bm{\theta}$. Once trained, the event detection model quickly adapts to unseen tasks even without examples~\cite{finn2017model, yue2023metaadapt}.

Given a set of tasks $\{ \bm{\mathcal{T}}_{i} \}_{i=1}^{M}$ and model $\bm{f}$ parameterized by $\bm{\theta}$, \ours aims at minimizing the overall evaluation loss of the tasks (as in \Cref{eq:objective}). \ours consists of an inner-loop optimization stage (i.e., $\mathcal{A}lg$) and an outer-loop optimization staget that minimizes the overall loss w.r.t. $\bm{\theta}$. For the inner-loop update, $\mathcal{A}lg$ denotes gradient descent with learning rate $\alpha$, i.e.:
\begin{equation}
    \label{eq:inner_lever}
    \mathcal{A}lg(\bm{\theta}, \bm{\mathcal{D}}^{\mathrm{train}}) = \bm{\theta} - \alpha \nabla_{\bm{\theta}} \mathcal{L}(\bm{\theta}, \bm{\mathcal{D}}^{\mathrm{train}}) = \bm{\phi},
\end{equation}
we denote the updated parameter set with $\bm{\phi}$. In the outer-level optimization, we are interested in learning an optimal set $\bm{\theta}$ that minimizes the meta loss on the evaluation sets. The learning is achieved by differentiating through the inner-loop optimization (i.e., $\mathcal{A}lg$) back to the initial parameter set $\bm{\theta}$, which requires the computation of second-order gradients or first-order approximation (as shown in~\Cref{fig:method}). Specifically, we derive the gradients w.r.t. $\bm{\theta}$:
\begin{equation}
\label{eq:analysis}
\frac{d \mathcal{L}}{d \bm{\theta}} = \frac{d \bm{\phi}}{d \bm{\theta}} \nabla_{\bm{\phi}} \mathcal{L}(\mathcal{A}lg(\bm{\theta}, \bm{\mathcal{D}}^{\mathrm{train}}), \bm{\mathcal{D}}^{\mathrm{test}}),
\end{equation}
notice that $\mathcal{A}lg(\bm{\theta}, \bm{\mathcal{D}}^{\mathrm{train}})$ is equivalent to $\bm{\phi}$. Component $\nabla_{\bm{\phi}} \mathcal{L}(\mathcal{A}lg(\bm{\theta}, \bm{\mathcal{D}}^{\mathrm{train}}), \bm{\mathcal{D}}^{\mathrm{test}})$ refers to first-order gradients w.r.t. the task-specific parameter set $\bm{\phi}$ (i.e., $\mathcal{L} \rightarrow \bm{\phi}$). The left component $\frac{d \bm{\phi}}{d \bm{\theta}}$ tracks parameter-to-parameter changes from $\bm{\phi}$ to $\bm{\theta}$ through $\mathcal{A}lg$ (i.e., $\bm{\phi} \rightarrow \bm{\theta}$), which involves the computation of the Hessian matrix. As the estimation of the matrix $\frac{d \bm{\phi}}{d \bm{\theta}}$ requires extensive computational resources, we provide both first-order and second-order implementations for \ours.

\noindent
\textbf{Zero-Shot \ours}: For the zero-shot evaluation, the learned initial parameter set should be directly evaluated on $\bm{\mathcal{D}}^{\mathrm{test}}$ for an unseen task. Therefore, directly optimizing \Cref{eq:objective} is not feasible for zero-shot event detection, as $\bm{\mathcal{D}}^{\mathrm{train}}$ is not provided. For optimization, however, training event types can be used for inner-loop optimization of the model. As such, we sample $\bm{\mathcal{D}}^{\mathrm{train}}$ and $\bm{\mathcal{D}}^{\mathrm{test}}$ from \emph{different training tasks} to improve the model generalization on unseen events. Specifically for training task $\bm{\mathcal{T}}_{j}$, we optimize
\begin{equation}
    \label{eq:zero-shot-objective}
    \min_{\substack{\bm{\theta}}} \mathcal{L}(\mathcal{A}lg(\bm{\theta}, \bm{\mathcal{D}}^{\mathrm{train}} \sim \{ \bm{\mathcal{T}}_{i} \}_{i=1, i \neq j}^{M}), \bm{\mathcal{D}}_{j}^{\mathrm{test}}),
\end{equation}
where $\bm{\mathcal{D}}^{\mathrm{train}}$ is a disjoint training set sampled from $\{ \bm{\mathcal{T}}_{i} \}_{i=1, i \neq j}^{M}$. As a result, the model `learns to adapt' to unseen events by optimizing on different training and evaluation sets. To improve the performance on unseen event types via Hungarian algorithm, we additionally design a contrastive loss term in the meta objective to learn class-separating features, which is introduced in \Cref{sec:meta_objective}.

\noindent
\textbf{Few-Shot \ours}: In the few-shot event detection, we directly sample training tasks and optimize the model as in \Cref{eq:objective}. Similar to the zero-shot \ours, the parameters are updated upon the tasks separately (i.e., $\bm{\phi}$) in each iteration based on the initial parameters $\bm{\theta}$. Then, the meta loss $\mathcal{L}$ and gradients w.r.t. $\bm{\theta}$ are computed for each task using the updated $\bm{\phi}$ and the evaluation sets. 
In our implementation, we adopt layer- and step-adaptive learning rates for inner-loop optimization and cosine annealing to improve the convergence of \ours~\cite{antoniou2018train}.

\subsection{Meta Objective}
\label{sec:meta_objective}
We now introduce our training objective $\mathcal{L}$ for \ours. Based on the proposed model in \Cref{sec:model_design}, our loss function contains two classification losses: trigger classification loss $\mathcal{L}_{\mathrm{trigger}}$ and event classification loss $\mathcal{L}_{\mathrm{event}}$ (i.e., negative log likelihood loss). To enlarge the inter-class event discrepancy for improved zero- and few-shot performance, we additionally propose a contrastive loss $\mathcal{L}_{\mathrm{con.}}$ based on the maximum mean discrepancy (MMD)~\cite{gretton2012kernel, yue-etal-2021-contrastive, yue-etal-2022-domain, yue2022contrastive}. In particular, we measure the discrepancy between two different event types by estimating the MMD distance. MMD computes the distance between two event distributions using an arbitrary number of input features drawn from these event types. Mathematically, MMD distance between input features $\bm{X}$ and $\bm{Y}$ can be computed as:
\begin{equation}
  \begin{aligned}
    \mathcal{D}(\bm{X}, \bm{Y}) &= \frac{1}{|\bm{X}||\bm{X}|} \sum_{i=1}^{|\bm{X}|} \sum_{j=1}^{|\bm{X}|} k(\psi(\bm{x}^{(i)}), \psi(\bm{x}^{(j)})) \\
    &+ \frac{1}{|\bm{Y}||\bm{Y}|} \sum_{i=1}^{|\bm{Y}|} \sum_{j=1}^{|\bm{Y}|} k(\psi(\bm{y}^{(i)}), \psi(\bm{y}^{(j)})) \\
    &- \frac{2}{|\bm{X}||\bm{Y}|} \sum_{i=1}^{|\bm{X}|} \sum_{j=1}^{|\bm{Y}|} k(\psi(\bm{x}^{(i)}), \psi(\bm{y}^{(j)})),
  \label{eq:mmd}
  \end{aligned}
\end{equation}
where $k$ represents the Gaussian kernel and $\psi$ represents the feature mapping function defined by the transformer network (i.e., feature encoder).

Based on the MMD distance, we further compute the inter-class distances for all pairs of event types. Suppose $\bm{X}_{i}$ represents the set of trigger-aware event features (i.e., concatenation of $[\bm{v};\bm{t}]$ as in \Cref{sec:model_design}) of the $i$-th class, the contrastive loss can be formulated for the $N$-way setting as:
\begin{equation}
    \mathcal{L}_{\mathrm{con.}} = - \frac{1}{N(N-1)} \sum_{i=1}^{N} \sum_{j=1, j \neq i}^{N} \mathcal{D}(\bm{X}_{i}, \bm{X}_{j}),
    \label{eq:contrastive_loss}
\end{equation}
in which we compute $N\times (N-1)$ pairs of inter-class distances, such inter-class distances are maximized (by taking the negative values) in the meta objective function to encourage class-separating features in \ours, and therefore improves both zero- and few-shot performance in event detection.

\noindent
\textbf{Overall Objective}: We now combine the mentioned terms into a single optimization objective for \ours in \Cref{eq:loss}. In our objective function, $\mathcal{L}_{\mathrm{event}}$ and $\mathcal{L}_{\mathrm{trigger}}$ represent the event and trigger classification loss (i.e., negative log likelihood loss), and $\mathcal{L}_{\mathrm{con.}}$ denotes the contrastive loss based on MMD. The overall objective contains three terms and $\mathcal{L}_{\mathrm{con.}}$ is weighted by a scaling factor $\lambda_{c}$ (to be chosen empirically):
\begin{equation}
    \mathcal{L} = \mathcal{L}_{\mathrm{event}} + \mathcal{L}_{\mathrm{trigger}} + \lambda_{\mathrm{c}} \mathcal{L}_{\mathrm{con.}}.
    \label{eq:loss}
\end{equation}

\subsection{Overall Framework}
The overall framework of \ours is presented in \Cref{fig:method}. The right subfigure illustrates the proposed model that integrates attentive trigger features for prompt-based event detection. The left subfigure illustrates the meta training paradigm of \ours, where the initial parameter $\bm{\theta}$ is updated and evaluated upon sampled tasks using the proposed contrastive loss in \Cref{eq:loss}. Then the gradients w.r.t. $\bm{\theta}$ can be computed (via second-order optimization or first-order approximation) to update the initial model. Unlike previous works~\cite{schick-schutze-2021-exploiting, cong-etal-2021-shot, li2022piled, zhang-etal-2022-zero}, we design a trigger-aware model for efficient training and inference on event detection tasks. Moreover, we propose a meta learning framework \ours with a fine-grained contrastive objective function for zero- and few-shot event detection, which encourages generalizable and class-separating features across both seen and unseen event types.

%% file: 4_experiment.tex
\section{Experiments}
\label{sec:experiment}

\begin{table*}[t]
\centering
\setlength\tabcolsep{3pt}
\begin{tabular}{@{}lcccccc@{}}
\toprule
\multirow{2}{*}{\textbf{Method}} & \multicolumn{3}{c}{\textbf{FewEvent}}                                                       & \multicolumn{3}{c}{\textbf{MAVEN}}                                                          \\ \cmidrule(l){2-7} 
                                 & \textbf{\, F1 $\uparrow$ \,} & \textbf{\, AMI $\uparrow$ \,} & \textbf{\, FM $\uparrow$ \,} & \textbf{\, F1 $\uparrow$ \,} & \textbf{\, AMI $\uparrow$ \,} & \textbf{\, FM $\uparrow$ \,} \\ \midrule
\textbf{SCCL}                    & 0.3184          & 0.2371          & 0.2436          & 0.2424          & 0.1546          & 0.1483          \\
\textbf{SS-VQ-VAE}               & 0.3670          & 0.3462          & 0.2758          & 0.1934          & 0.1192          & \ul{0.1838}     \\
\textbf{BERT-OCL}                & 0.3296          & \ul{0.5326}     & 0.4016          & 0.1446          & \ul{0.1915}     & 0.1160          \\
\textbf{ZEOP}                    & 0.4869          & 0.4065          & 0.3392          & \ul{0.2444}     & 0.1274          & 0.1642          \\
\textbf{ZEOP*}                   & \ul{0.5655}     & 0.5135          & \ul{0.4360}     & 0.2383          & 0.1366          & 0.1484          \\
\textbf{\ours}                   & \textbf{0.6837}$_{\small{\pm\text{0.0689}}}$ & \textbf{0.6884}$_{\small{\pm\text{0.0315}}}$ & \textbf{0.7247}$_{\small{\pm\text{0.0807}}}$ & \textbf{0.3686}$_{\small{\pm\text{0.0412}}}$ & \textbf{0.2352}$_{\small{\pm\text{0.0521}}}$ & \textbf{0.2569}$_{\small{\pm\text{0.0392}}}$     \\ \bottomrule
\end{tabular}
\caption{Zero-Shot event detection results ($10$-way for both datasets).}
\label{tab:zero-shot}
\vspace{-10pt}
\end{table*}

\subsection{Settings}
\noindent
\textbf{Model}: Following previous work~\cite{wang-etal-2021-cleve, li2022piled}, we select RoBERTa as the base model in \ours~\cite{liu2019roberta}.

\noindent
\textbf{Evaluation}: To validate the proposed method, we follow~\cite{chen-etal-2021-honey, cong-etal-2021-shot, li2022piled} to split the datasets into train, validation and test sets. For evaluation metrics, we adopt micro F1 score as main performance indicator. For the zero-shot setting, we additionally adopt adjusted mutual information (AMI) and Fowlkes Mallows score (FM) to evaluate clustering performance. See evaluation details in \Cref{sec:implementation}.

\noindent
\textbf{Datasets and Baselines}: To examine \ours performance, we adopt publicly available datasets FewEvent and MAVEN~\cite{deng2020meta, wang-etal-2020-maven} and state-of-the-art baseline methods for comparison. For zero-shot baselines, we adopt SCCL~\cite{zhang-etal-2021-supporting}, SS-VQ-VAE~\cite{huang-ji-2020-semi}, BERT-OCL~\cite{zhang-etal-2022-zero} and ZEOP~\cite{zhang-etal-2022-zero}. For the few-shot setting, we choose BERT-CRF~\cite{devlin-etal-2019-bert}, PA-CRF~\cite{cong-etal-2021-shot}, Prompt+QA~\cite{li2022piled} and P4E~\cite{li2022piled} as baselines\footnote{We additionally select the adapted ZEOP*, PA-CRF* and P4E* as improved variants of the baselines, see \Cref{sec:baselines}.}. Dataset and baseline details are in \Cref{sec:implementation}

\noindent
\textbf{Implementation}: We use the \texttt{roberta-base} variant in our implementation, our default model is trained with AdamW optimizer with zero weight decay and cosine annealing for meta learning rate of $1e-5$. For inner-loop optimization, we use layer- and step-adaptive learning rates with an intial learning rate of $1e-3$, where the model is updated $50$ times in each task. We select the best model on the validation set for evaluation on the test set. For baseline methods, we follow the reported training procedure and hyperparameter settings from the original works unless otherwise suggested. More implementation and experiment details are provided in \Cref{sec:implementation}.

\subsection{Zero-Shot Results}
We first report zero-shot results on all datasets in \Cref{tab:zero-shot}, which is divided into two parts by the used datasets. We perform $10$-way event detection on unseen tasks from the disjoint test sets and report the evaluation results, the best results are marked bold, the second best results are underlined. We observe: 
(1)~the zero-shot performance on FewEvent is comparatively higher than MAVEN for both baselines and \ours, possibly due to the increased coverage of event domains in MAVEN. (2)~\ours performs the best on both datasets, in particular, \ours achieves $35.9\%$ average improvements on F1 compared to the second best-performing method. (3)~Despite lower performance on MAVEN, \ours outperforms all baseline methods with up to $50.8\%$ improvements on F1, suggesting the effectiveness of the proposed meta training algorithm in detecting unseen events.

\begin{figure}[t]
    \centering
    \includegraphics[trim=0 0 0.5cm 0, clip, width=0.9\linewidth]{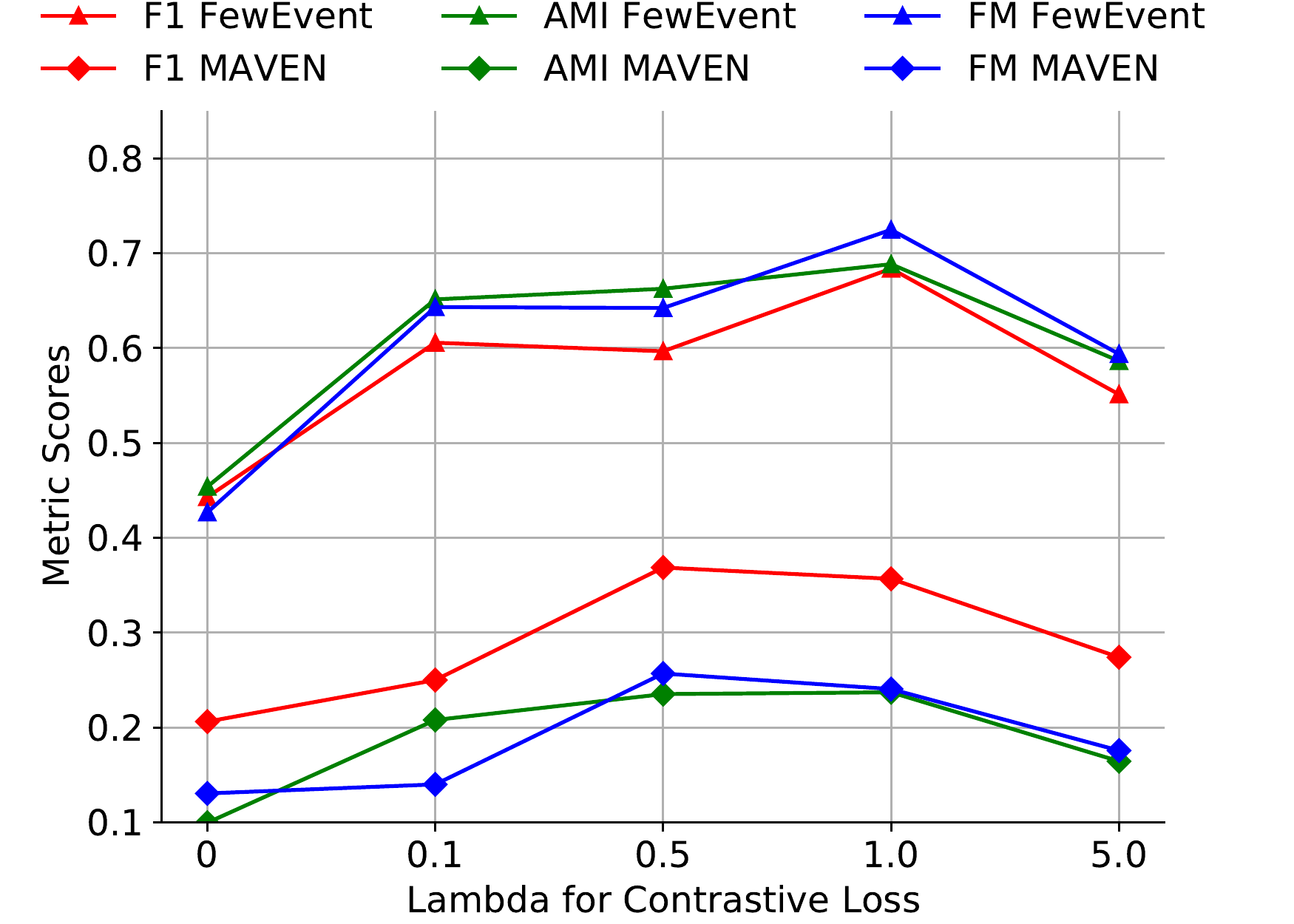}
    \caption{Sensitivity analysis of $\lambda_{c}$.}
    \label{fig:sensitivity_contrastive}
    \vspace{-10pt}
\end{figure}

We further study the effectiveness of the proposed contrastive loss quantitatively in zero-shot experiments by varying scaling factor $\lambda_{c}$. Specifically, we choose $\lambda_{c}$ from $0$ to $5$ and evaluate the performance changes on both datasets. The results are visually presented in \Cref{fig:sensitivity_contrastive}, from which we observe: (1)~by applying the contrastive loss (i.e., $\lambda_{c} \neq 0$) in the meta objective, the zero-shot performance consistently improves regardless of the choice of $\lambda_{c}$; (2)~despite huge improvements, carefully chosen $\lambda_{c}$ is required for the best zero-shot performance (up to $\sim 100\%$ increases across all metrics). Overall, the contrastive objective proposed in \ours is particularly effective in learning class-separating features, and thereby improves the zero-shot event detection performance.

\begin{figure}[t]
    \centering
    \includegraphics[trim=7.2cm 4.3cm 7cm 4.3cm, clip, width=1.0\linewidth]{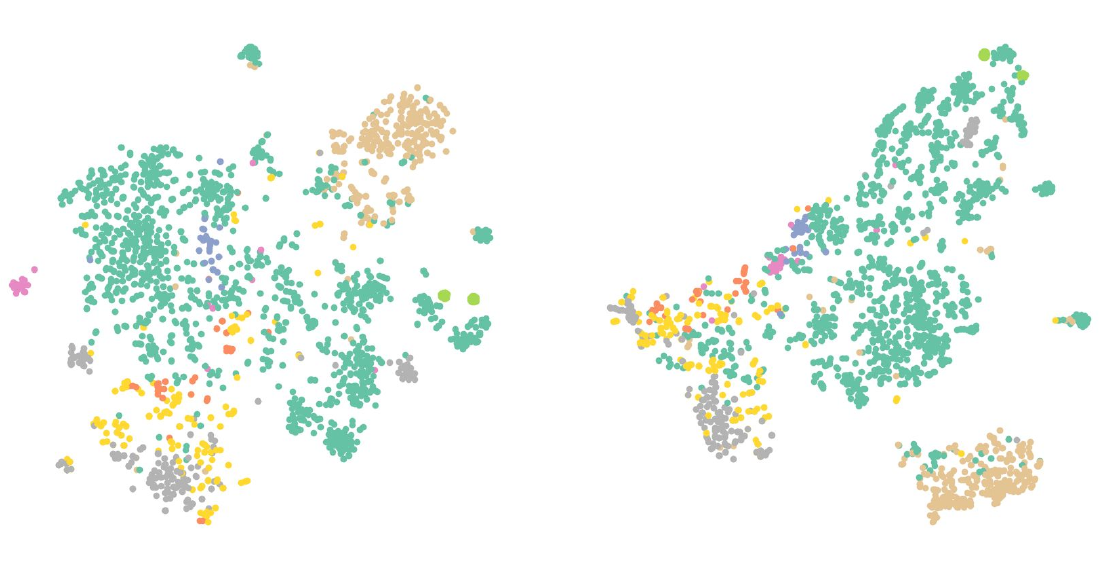}
    \caption{Feature visualization of \ours w/o (left subfigure) and w/ (right subfigure) contrastive learning.}
    \label{fig:qualitative_analysis}
    \vspace{-10pt}
\end{figure}

\begin{table*}[t]
\centering
\setlength\tabcolsep{12pt}
\begin{tabular}{@{}lcccc@{}}
\toprule
\multirow{2}{*}{\textbf{Method}} & \multicolumn{2}{c}{\textbf{FewEvent}}                        & \multicolumn{2}{c}{\textbf{MAVEN}}                           \\ \cmidrule(l){2-5} 
                                 & \textbf{\, F1 (K=5) $\uparrow$ \,} & \textbf{\, F1 (K=10) $\uparrow$ \,} & \textbf{\, F1 (K=5) $\uparrow$ \,} & \textbf{\, F1 (K=10) $\uparrow$ \,} \\ \midrule
\textbf{BERT-CRF}                & 0.4406                       & 0.6673                        & 0.4814                       & 0.6468                        \\
\textbf{PA-CRF}                  & 0.5848                       & 0.6164                        & 0.4257                       & 0.4918                        \\
\textbf{PA-CRF*}                 & 0.6364                       & 0.7069                        & 0.5316                       & 0.6562                        \\
\textbf{Prompt + QA}             & 0.6523                       & 0.6750                        & 0.4786                       & 0.6543                        \\
\textbf{P4E}                     & 0.8198                       & 0.8550                        & 0.6064                       & 0.6951                        \\
\textbf{P4E*}                    & \ul{0.9070$_{\small{\pm\text{0.0220}}}$} & \ul{0.9270$_{\small{\pm\text{0.0110}}}$} & \ul{0.6390$_{\small{\pm\text{0.0090}}}$} & \ul{0.7260$_{\small{\pm\text{0.0150}}}$} \\
\textbf{\ours \quad \quad \quad} & \textbf{0.9318}$_{\small{\pm\text{0.0216}}}$ & \textbf{0.9576}$_{\small{\pm\text{0.0052}}}$ & \textbf{0.9306}$_{\small{\pm\text{0.0026}}}$ & \textbf{0.9486}$_{\small{\pm\text{0.0003}}}$ \\ \bottomrule
\end{tabular}
\caption{Few-shot event detection results ($10$-way for FewEvent and $45$-way for MAVEN).}
\label{tab:few-shot}
\vspace{-10pt}
\end{table*}

We additionally present qualitative analysis of our zero-shot results by comparing the feature visualization (via T-SNE) with and without the proposed contrastive loss. We use color to represent different event types and present the visualization in \Cref{fig:qualitative_analysis}. With the contrastive loss (right subfigure), the model generates class-separating features on unseen events compared to \ours trained without contrastive loss (left subfigure). Examples of the same event type are also more aggregated, showing improved clustering results, which can be combined with trigger predictions for identifying unseen event types. In sum, the proposed contrastive loss demonstrates effectiveness in zero-shot settings and consistently outperforms baselines.

\begin{table}[t]
\centering
\begin{tabular}{@{}lcc@{}}
\toprule
\textbf{Method}     & \textbf{F1 (K=5) $\uparrow$} & \textbf{F1 (K=10) $\uparrow$} \\ \midrule
\ours               & 0.9318                       & 0.9576 \\
\, w/o Trigger      & 0.9170                       & 0.9367 \\
\, w/o Verbalizer   & 0.8117                       & 0.8516 \\
\, w/o Meta Learner & 0.6257                       & 0.6390 \\ \bottomrule
\end{tabular}
\caption{Ablation study of \ours.}
\label{tab:ablation}
\vspace{-10pt}
\end{table}

\subsection{Few-Shot Results}
To examine the effectiveness of our method for both zero- and few-shot scenarios, we also perform few-shot experiments on both datasets. The $5$-shot and $10$-shot event detection experiment results are presented in \Cref{tab:few-shot}, where all evaluation event types are used ($10$-way for FewEvent and $45$-way for MAVEN\footnote{Similar to~\cite{chen-etal-2021-honey}, we perform binary classification for each of the event types in MAVEN as multiple event labels may exist on the same input context.}). We observe: (1)~all baseline methods and \ours achieve improved performance in few-shot event detection compared to the zero-shot results. For example, \ours achieves $36.3\%$ F1 improvement in $5$-shot setting on FewEvent. (2)~For MAVEN, the average performance of all baseline methods are comparatively lower than FewEvent, indicating the challenge of few-shot classification with increased number of event types. (3)~\ours achieves the best results, outperforming the second best method in F1 by $3.0\%$ (on FewEvent) and $38.1\%$ (on MAVEN) on average across both few-shot settings. Overall, \ours achieves state-of-the-art performance in event detection even only with few examples.

We now perform ablation studies in the few-shot setting to evaluate the effectiveness of the proposed component in \ours.
In particular, we remove the proposed attentive trigger features (i.e., trigger), trigger-aware soft verbalizer (i.e., verbalizer) and outer-loop optimization (i.e., meta learner) in sequence to observe the performance changes on FewEvent. The results are reported in \Cref{tab:ablation}. For all components, we observe performance drops when removed from \ours. In the $5$-shot setting, F1 score reduces by $1.6\%$ and $12.9\%$ respetively when removing the attentive trigger features and trigger-aware soft verbalizer. The results suggest that proposed components are effective for improving few-shot event detection.

\begin{table}[t]
\centering
\begin{tabular}{@{}lcc@{}}
\toprule
\textbf{Method}     & \textbf{F1 (K=5) $\uparrow$} & \textbf{F1 (K=10) $\uparrow$} \\ \midrule
Prompt A            & 0.9318                       & 0.9576 \\
Prompt B            & 0.9363                       & 0.9644 \\
Prompt C            & 0.9272                       & 0.9527 \\
Prompt D            & 0.9236                       & 0.9592 \\ \bottomrule
\end{tabular}
\caption{Analysis of different prompt design.}
\label{tab:prompt}
\vspace{-10pt}
\end{table}

Finally, we study the influence of prompt designs and report the results on FewEvent in \Cref{tab:prompt}. In particular, we select from prompt A: `\texttt{A <mask> event}', B: `\texttt{This text describes a <mask> event}', C: `\texttt{This topic is about <mask>}' and D: `\texttt{[Event: <mask>]}'. From the results we observe: for $5$-shot event detection, prompt A and B perform the best while prompt B and D achieves better performance with $0.9644$ and $0.9592$ F1 in $10$-shot setting. On average, prompt B outperforms all other prompt designs in the F1 metric, indicating that well-designed instructions may slightly improve few-shot event detection results.

%% file: 5_conclusion.tex
\section{Conclusion}
In this paper, we design a meta learning framework \ours for zero- and few-shot event detection. \ours proposes to leverage attentive trigger features for efficient inference and predicts via a trigger-aware soft verbalizer. Moreover, the proposed \ours trains the model to search for the optimal parameter set for quick adaptation to unseen event detection tasks. Extensive experiment results demonstrate the effectiveness of \ours by consistently outperforming state-of-the-art methods on benchmark datasets in both zero- and few-shot event detection.

\section{Limitations}
While the proposed \ours achieves significant improvements in both zero- and few-shot event detection, \ours requires additional computational resources due to the layer- and step-adaptive learning rates and the outer-loop optimization, which may cause increased computational costs for training \ours. Moreover, we have not investigated the benefits of task scheduling techniques and similarity-based meta learning in \ours to fully explore the training event types. Consequently, we plan to study efficient meta learning with advanced training task scheduling for further improvements in zero- and few-shot event detection as future work.

%% file: 6_appendix.tex
\section{Implementation}
\label{sec:implementation}

\subsection{Datasets}
\label{sec:datasets}

We adopt FewEvent and MAVEN for our experiments, the details of the datasets are reported below.

\noindent
\textbf{FewEvent} is a dataset designed for few-shot event detection~\cite{deng2020meta}. We follow the preprocessing of~\cite{cong-etal-2021-shot, li2022piled} and present the data statistics in \Cref{tab:fewevent}. FewEvent contains $100$ event types with three disjoint sets of event classes in training, validation and test sets. The dataset is based on ACE and TAC-KBP with new event types from Freebase and Wikipedia.

\begin{table}[h]
\centering
\begin{tabular}{@{}l|ccc@{}}
\toprule
         & Train & Valid & Test \\ \midrule
Classes  & 80    & 10    & 10   \\
Examples & 68506 & 2173  & 697  \\ \bottomrule
\end{tabular}
\caption{Dataset statistics of FewEvent.}
\label{tab:fewevent}
\end{table}

\noindent
\textbf{MAVEN} is a large event detection dataset with over $150$ event types and over $80$k event mentions in total~\cite{wang-etal-2020-maven}. We follow the preprocessing of~\cite{chen-etal-2021-honey, li2022piled} and present the data statistics in \Cref{tab:maven}. MAVEN covers an enlarged set of event types with increased examples per class. Unlike FewEvent, the event types in the validation and test sets are overlapping. Since MAVEN provide multi-label examples, we perform binary classification for each of the event types, we additionally sample $10$ times the negative examples for both training and evaluation.

\begin{table}[h]
\centering
\begin{tabular}{@{}l|ccc@{}}
\toprule
         & Train  & Valid & Test \\ \midrule
Classes  & 125    & 45    & 45   \\
Examples & 79906  & 1532  & 1555 \\ \bottomrule
\end{tabular}
\caption{Dataset statistics of MAVEN.}
\label{tab:maven}
\end{table}

\subsection{Baseline Methods}
\label{sec:baselines}

We introduce the details of the zero-shot baseline methods, followed by the baseline methods in the few-shot setting. For zero-shot methods that leverage unseen event examples in training (e.g., ZEOP), we adapt the baseline methods by dividing the training set into seen and unseen event types. As such, unseen examples can be sampled from the training set and no examples from the evaluation event types are participated in training. 

\textbf{Supporting Clustering with Contrastive Learning (SCCL)} is a clustering-based approach for unsupervised classification. SCCL is used to detect new event types based on unseen event mentions. The contextual feature of trigger tokens are used in our experiments~\cite{zhang-etal-2021-supporting}.

\textbf{Semi-supervised Vector Quantized Variational Autoeocoder (SS-VQ-VAE)} leverages variational autoencoder to learn discrete event features. SS-VQ-VAE is trained on seen event types and annotations and can be adapted for zero-shot event detection~\cite{huang-ji-2020-semi}.

\textbf{BERT Ordered Contrastive Learning (BERT-OCL)} designs an ordered contrastive learning method for clustering unseen event types. The Euclidean distance is used to compute pair-wise distance between examples for reducing intra-class distances and increasing inter-class distances~\cite{devlin-etal-2019-bert, zhang-etal-2022-zero}.

\textbf{Zero-Shot Event Detection with Ordered Contrastive Learning (ZEOP \& ZEOP*)} leverages prompt learning and ordered contrastive loss based on both instance-level and class-level distance for zero-shot event detection. ZEOP first identifies trigger tokens then predicts event types by clustering, while ZEOP* predicts without inference on trigger words~\cite{zhang-etal-2022-zero}.

The following methods are the few-shot baseline methods used in our experiments.

\textbf{BERT Conditional Random Field (BERT-CRF)} uses BERT encoder with a conditional random field classifier used to classify tokens. BERT-CRF can be fine-tuned on event detection tasks with limited examples~\cite{devlin-etal-2019-bert}.

\textbf{Prototypical Amortized Conditional Random Field (PA-CRF \& PA-CRF*)} improves upon BERT-CRF by estimating transition scores and class uncertainty based on label prototypes and Gaussian distributions. PA-CRF* memories the prototypes and recomputes them in each iteration to achieve improved performance~\cite{cong-etal-2021-shot}.

\textbf{Prompt + Question Answering (Prompt+QA)} leverages both prompt-based classification and question answering to: (1)~make inference on the event type with predefined prompt and a verbalizer; (2)~perform QA task to query the trigger tokens based on the previous classification results~\cite{du-cardie-2020-event, li2022piled}.

\textbf{Prompt-Guided Event Detection (P4E \& P4E*)} proposed a prompt-based approach to first identify event types, followed by trigger localization using the previous output. P4E achieves state-of-the-art performance by prompt-tuning pretrained language models on event detection tasks. P4E* only performs event type classification without inference on trigger words~\cite{li2022piled}.

\subsection{Implementation Details}
For our evaluation method, we follow the previous works~\cite{chen-etal-2021-honey, cong-etal-2021-shot, li2022piled, zhang-etal-2022-zero} and split the datasets into train, validation, and test sets. The validation sets are used for selecting the best model (with validation F1 score) to perform evaluation. For baseline implementation, we follow the reported training procedure and hyperparameter settings from the original works unless otherwise suggested. However, for baseline methods that require unlabeled examples from unseen classes in training (e.g., ZEOP), we modify such methods by sampling event types from the training set as unseen events. As such, the baseline methods can be trained without test event types being participated in the optimization. As a result, we observe slight performance drops compared to the original implementation~\cite{zhang-etal-2022-zero}. For few-shot event detection baseline methods, the results are directly taken from~\cite{li2022piled}.

In \ours optimization, the outer-loop learning rates are selected from $[1e-5, 2e-5, 3e-5]$, the initial inner-loop learning rates are selected from $[1e-4, 1e-3]$\footnote{For inner-loop optimization, the verbalizer weights are initialized with $10$ times the base learning rate (i.e., inner-loop learning rate) for faster convergence.}, the learning rate of adaptive learning rate is $1e-4$. \ours uses the AdamW optimizer without cosine annealing learning rate scheduler in meta optimization. For inner-loop, the maximum batch size is $50$ and we leverage per-layer per-step adaptive learning rates and perform $50$ updates in total~\cite{antoniou2018train}. We adopt $[2, 3]$ as the number of tasks, number of iterations are selected from $[250, 500]$ depending on the task and size of the dataset. The model is validated every $25$ iterations, the best model in validation is used to evaluate on the test split. For hyperparameter, see sensitivity analysis in \Cref{sec:experiment}. All reported results are based on first-order meta learning approximation and experiments are performed on multiple NVIDIA A40 GPUs.

\begin{table}[t]
\centering
\begin{tabular}{@{}lcc@{}}
\toprule
\textbf{Metric}     & \textbf{FewEvent} & \textbf{MAVEN} \\ \midrule
Reported F1         & 0.6837            & 0.3686 \\
Rand Score          & 0.9164            & 0.8387 \\
Adjusted Rand       & 0.6780            & 0.2001 \\
Normalized MI       & 0.6719            & 0.3248 \\
Homogeneity Score   & 0.6779            & 0.3305 \\ \bottomrule
\end{tabular}
\caption{Additional zero-shot results of \ours.}
\label{tab:additional_zero_shot}
\vspace{-10pt}
\end{table}

\section{Additional Results}
\label{sec:additional_result}

In this section, we present additional results on zero-shot performance of \ours. Specifically, we provide additional clustering metrics for \ours on both datasets, with the results presented in \Cref{tab:additional_zero_shot}. We adopt the following clustering metrics: (1)~rand score (Rand Score); (2)~adjusted rand score (Adjusted Rand); (3)~normalized mutual information (Normalized MI) and (4)~homegeneity score (Homogeneity Score). Surprisingly, the rand score and adjusted rand score demonstrates significant difference on MAVEN, suggesting disproportionate label distribution in MAVEN. For the rest metrics on both datasets, we observe similar magnitude of performance as reported in \Cref{sec:experiment}. Moreover, we present additional visualization results on FewEvent below with varying $\lambda_{c}$ values. Compared to \Cref{fig:qualitative_analysis}, we observe that cluster aggregation worsens with reduced $\lambda_{c}$ values.

\begin{figure}[h]
    \centering
    \includegraphics[trim=1cm 3.5cm 1cm 3.5cm, clip, width=0.75\linewidth]{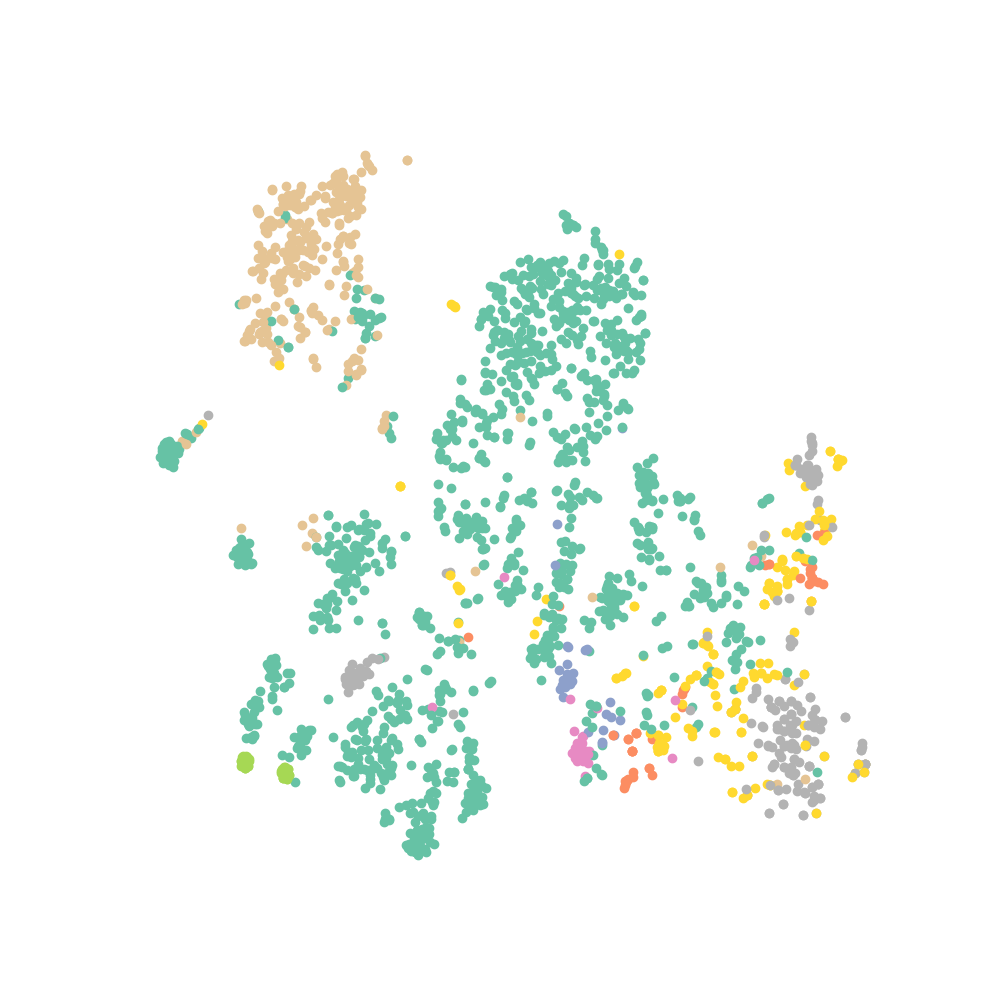}
    \caption{Feature visualization ($\lambda_{c} = 0.1$).}
    \vspace{-10pt}
\end{figure}

\begin{figure}[h]
    \centering
    \includegraphics[trim=1cm 3.5cm 1cm 3.5cm, clip, width=0.75\linewidth]{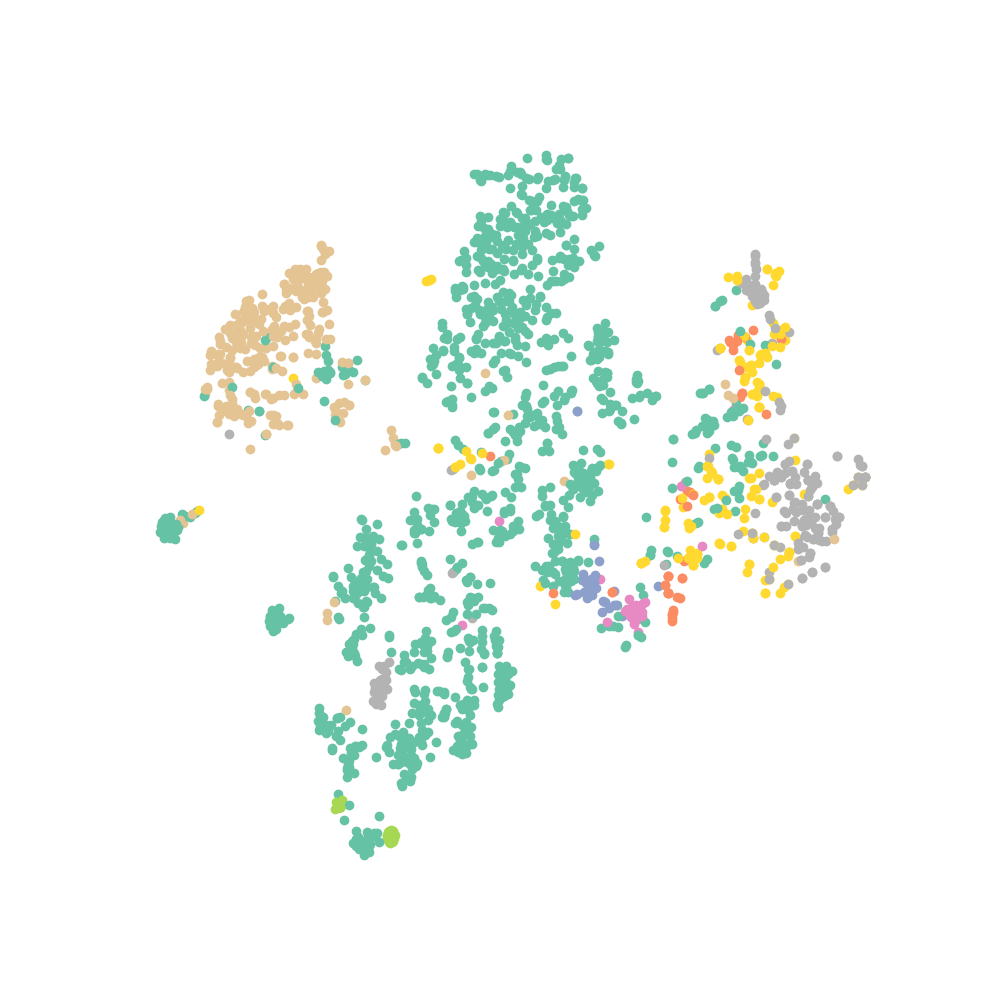}
    \caption{Feature visualization ($\lambda_{c} = 0.5$).}
    \vspace{-10pt}
\end{figure}